\documentclass[a4paper, oneside, twocolumn, notitlepage, 10pt]{extarticle_ecoc}
\usepackage{ecoc}

\usepackage{mathtools}
\def\ve#1{{\mathchoice{\mbox{\boldmath$\displaystyle #1$}}%
{\mbox{\boldmath$\textstyle #1$}}%
{\mbox{\boldmath$\scriptstyle #1$}}%
{\mbox{\boldmath$\scriptscriptstyle #1$}}}}

\def\Nmem{\mu}
\newcommand{\Dkl}{\mathrm{D_{KL}}}

\newcommand{\argmaxst}[3]{%
  &\arg\max_{#1} #2 \nonumber\\
  &\text{subject to} \quad #3
}

\begin{document}
\selectlanguage{english}    


\title{Sequential Neural Probabilistic Amplitude Shaping: Learning the Channel’s Language}%


\author{
    Mohammad Taha Askari\textsuperscript{(1,2)}, Lutz Lampe\textsuperscript{(1)},
    Amirhossein Ghazisaeidi\textsuperscript{(2)}
}

\maketitle                  


\begin{strip}
    \begin{author_descr}

        \textsuperscript{(1)} Department of Electrical and Computer Engineering, University of British Columbia, Vancouver, BC V6T 1Z4, Canada,
        \textcolor{blue}{\uline{mohammadtaha@ece.ubc.ca}} 
        \textsuperscript{(2)} Nokia Bell Labs, 12 rue Jean Bart, 91300 Massy, France

    \end{author_descr}
\end{strip}

\renewcommand\footnotemark{}
\renewcommand\footnoterule{}


\begin{strip}
\begin{ecoc_abstract}
We present the first neural probabilistic amplitude shaping that outperforms existing methods while accounting for all implementation losses, using a block‑less, easily implementable sequential autoregressive encoder compatible with arithmetic distribution matching, yielding reduced rate loss and higher achievable information rates. \textcopyright2026 The Author(s)
\end{ecoc_abstract}
\end{strip}

\section{Introduction}
\label{sec:intro}

Probabilistic amplitude shaping (PAS) combines probabilistic shaping (PS) with forward error correction (FEC), while enabling flexible rate adaptation \cite{bocherer2015bandwidth}. In PAS, a distribution matcher maps information bits to amplitudes following a target distribution, while sign bits are set by the FEC encoder.
%
For nonlinear fiber channels, PAS yields both linear and nonlinear shaping gains \cite{dar2014shaping,askari2024probabilistic}. Finite-length distribution matchers, such as enumerative sphere shaping (ESS), can reduce nonlinear interference noise (NLIN). Shorter shaping blocklengths often increase nonlinear gains but also raise rate loss, creating a tradeoff between shaping efficiency and nonlinear tolerance \cite{amari2019introducing,askari2023probabilistic,fehenberger2020analysis}.

Recent work shows that sequences with identical marginal symbol distributions can experience different nonlinear distortion depending on temporal ordering  \cite{secondini2022new}. This indicates that performance on channels with memory depends on the joint symbol distribution. Sequence-selection methods exploit this by generating multiple candidates and selecting the one with the lowest predicted nonlinear distortion using a nonlinearity-aware metric \cite{civelli2024sequence,askari2024probabilistic}. Although effective, such rejection sampling requires repeated metric evaluations, often ignores inter-block effects, incurs additional rate loss due to candidate-index side-information signaling, and typically requires large candidate pools, resulting in high computational complexity \cite{civelli2024cost}.
%

Neural shaping is a new paradigm for directly optimizing joint symbol distributions. In this framework, a neural encoder specifies conditional symbol probabilities, while arithmetic distribution matching (ADM) maps information bits into sequences that follow the learned distribution \cite{baur2015arithmetic,askari2026neural}. The encoder is trained end‑to‑end with differentiable sampling and information-theoretic objectives targeting achievable information rate (AIR) \cite{askari2025neural, ait2020joint}.
%
%
Building on this idea, neural PS (NPS) and neural PAS (NPAS) employ recurrent neural networks to model joint distributions over fixed-length blocks of signed and unsigned symbols, respectively \cite{askari2025neural,askari2026neural}.
These blockwise methods learn intra‑block temporal dependencies and have demonstrated nonlinear gains over conventional PAS and sequence-selection schemes. However, their block formulation confines dependencies to a finite horizon, ignores correlations across adjacent blocks, and induces non‑stationary symbol statistics as the conditioning context grows within each block.
%
Moreover, prior formulations of NPS and NPAS do not explicitly account for the rate loss caused by learned symbol dependencies during training. As a result, highly expressive temporal structures can incur excessive rate loss that diminishes or even negates the AIR gains obtained from improved nonlinear tolerance.

In this work, we revisit neural shaping with a rate-loss-aware formulation. We derive a training objective that explicitly incorporates implementation rate loss and a tractable lower bound suitable for gradient-based optimization enabling learned joint distributions that balance nonlinear shaping gain and rate-loss efficiency. We introduce sequential NPAS (Seq-NPAS), a next-token prediction architecture that models symbol distribution conditioned on a fixed-length context. This sequential, block-less design naturally matches ADM, yields stationary symbol statistics, removes artificial block boundaries, and simplifies implementation.
Through numerical results, we show that explicitly accounting for rate loss is essential to realize net gains from joint-distribution learning. For the first time, we present a neural shaping approach that outperforms existing methods, while accounting for all implementation losses. 
Our shaping approach extends beyond the fiber Kerr nonlinearity, demonstrating broader applicability to nonlinear channels with memory.

\section{Problem Statement}
\label{sec:problem}

We consider a PAS-based coherent optical system employing an $M$-QAM constellation. Let $\ve{x}=(x_1,\ldots,x_N)$ and $\ve{y}=(y_1,\ldots,y_N)$ denote the transmitted and received symbol sequences, respectively. Each transmitted symbol is written as $x_t=(s_t,a_t)\in\{c_1,\ldots,c_M\}$, where $a_t$ is the unsigned QAM symbol and $s_t$ denotes the sign quadrant. The corresponding binary label is $\ve{b}_t=(b_t^{(1)},\ldots,b_t^{(m)})$, where $m=\log_2 M$.

For fixed channel parameters and launch power, the objective of PS is to optimize the input joint distribution $p(\ve{x})$ to maximize the AIR. Under bit-metric decoding and incorporating the rate loss, this can be expressed as \cite{ait2020joint}
\vspace{-2mm}
\begin{align}
\label{eq:BMD_rate}
\argmaxst{p(\ve{x})}
{\frac{1}{N}\sum_{t=1}^{N}
\bigg[
\underbrace{
H(\ve{b}_t)-\sum_{i=1}^{m}H\!\left(b_t^{(i)}\mid \ve{y}\right)
}_{R_t}
\bigg]
-R_{\mathrm{loss}}}
{\sum_{\ve{x}} p(\ve{x})\|\ve{x}\|^2 = 1,}
\end{align}
where $H(\cdot)$ denotes entropy,  $R_t$ is the instantaneous bit-metric rate, and
\vspace{-2mm}
\begin{equation}
\label{eq:rateloss}
R_{\mathrm{loss}}
=
\frac{1}{N}\sum_{t=1}^{N} H(\ve{b}_t)
-
\frac{1}{N}H(\ve{b}_1,\ldots,\ve{b}_N)
\end{equation}
is the rate loss induced by symbol dependencies. This term equals the average gap between the sum of marginal entropies and the entropy rate of the joint process. It therefore represents the minimum rate loss of an ideal distribution matcher and depends only on the learned joint distribution.

\section{Training Objective}
To solve \eqref{eq:BMD_rate}, we follow the neural shaping framework of \cite{askari2025neural}. The distribution $p(\ve{x})$ is parameterized by a neural network and sampled during training using the Gumbel-Softmax trick \cite{jang2016categorical} with a straight-through estimator \cite{bengio2013estimating}. The generated symbols are transmitted through a differentiable approximation of the channel,
and the received samples are processed by a mismatched Gaussian demapper that outputs log-likelihood ratios (LLRs).

In previous work \cite{askari2025neural,askari2026neural}, training is based on an adjusted binary cross entropy loss between LLRs and bit labels, which can be decomposed as
\vspace{-2mm}
\begin{equation}
\label{eq:loss_hat}
\begin{split}
&\mathcal{L}
=
\frac{1}{N} \sum_{t=1}^{N} \Big( - R_t \\
    &\quad + \sum_{i=1}^{m} \mathbb{E} \left[ \Dkl(p(b_t^{(i)} \mid \ve{y}) \,\|\, \tilde{p}(b_t^{(i)} \mid y_t)) \right] \Big),
\end{split}
\end{equation}
where $\tilde{p}(b_t^{(i)}\mid y_t)$ denotes the approximate bit posterior and $\Dkl$ is the Kullback–Leibler (KL) divergence capturing demapper mismatch.
%
To explicitly account for symbol dependencies, we propose the rate-loss-aware objective
\begin{equation}
\label{eq:loss_plus}
\mathcal{L}^{++}
=
\mathcal{{L}}
+
R_{\mathrm{loss}}
+
\lambda \,
\Dkl\!\left(
p(x)\,\|\,p_{\mathrm{MB}}(x)
\right),
\end{equation}
where $p(x)$ is the marginal symbol distribution induced by $p(\ve{x})$, $p_{\mathrm{MB}}(x)$ is a Maxwell--Boltzmann target distribution optimized for the channel, and $\lambda$ controls the tradeoff between marginal shaping and joint-distribution learning. The regularization term encourages the model to preserve a favorable marginal distribution while using temporal dependencies to improve performance.



\section{Sequential Neural Encoder}
\label{sec:seq}

We introduce the new Seq-NPAS in which the transmitted unsigned symbol process is modeled as a stationary autoregressive source with finite memory. It is based on writing the joint distribution of the unsigned symbols $\ve{a}$ as
\vspace{-1mm}
\begin{equation}
p_{\theta}(\ve{a})
=
p_{\ve{\theta}}\!\left(a_1^{\Nmem-1}\right)
\prod_{t=\Nmem}^{N}
p_{\ve{\theta}}\!\left(
a_t \mid a_{t-\Nmem+1}^{t-1}
\right),
\end{equation}
where $\Nmem$ denotes the conditioning memory and $\ve{\theta}$ represents the model parameters.
At each time step, the encoder predicts the next-symbol distribution from previously generated symbols, analogous to autoregressive language models that predict the next token from prior context. Here, however, the token alphabet is the constellation and the learned statistics are optimized for communication performance.


Unlike blockwise neural shaping methods such as NPAS, which learn independent finite-length block distributions and therefore cannot capture cross-block correlations, the proposed sequential model removes artificial block boundaries and applies the same time-invariant prediction rule at every symbol position. This avoids the position-dependent context of blockwise schemes, yields stationary symbol statistics, and enables consistent exploitation of temporal dependencies over long sequences when $\Nmem$ is matched to the effective channel memory.



\section{Numerical Results}
\label{sec:results}

We evaluate the 
methods in a dual-polarization wavelength-division multiplexed (WDM) system with $5$ channels operating at $50$~GBaud and $55$~GHz channel spacing. Symbols are pulse shaped using a root-raised-cosine filter with roll-off factor $0.1$. The transmission link consists of a single $205$~km span of standard single-mode fiber with attenuation $0.2$~dB/km, chromatic dispersion (CD) $17$~ps/nm/km, and nonlinear coefficient $1.3$~W$^{-1}$km$^{-1}$. Span loss is compensated by an erbium-doped fiber amplifier with $5$~dB noise figure. At the receiver, CD compensation and linear pilot-aided carrier phase recovery \cite{neshaastegaran2019log} with $1\%$ pilot overhead are applied. Reported results correspond to the central WDM channel.

As a blockwise baseline, we consider NPAS parameterized by a single-layer long short-term memory (LSTM) \cite{hochreiter1997long} network with hidden dimension $256$, followed by a linear projection to unsigned-symbol logits. The proposed Seq-NPAS employs a single-layer Transformer \cite{vaswani2017attention} with $8$ attention heads, hidden dimension $64$, and feed-forward dimension $256$. Rotary positional embeddings \cite{su2024roformer} are used to capture relative temporal structure, while 
Swish-gated linear unit activations improve training stability and parameter efficiency \cite{shazeer2020glu}.

All neural encoders are trained at their respective optimal launch power using 
a single-channel single-polarization perturbation-based channel from \cite{askari2026neural}.
During evaluation, the trained models are implemented through ADM to generate symbol sequences, which are then transmitted over the dual-polarization WDM link simulated using the split-step Fourier method across multiple launch powers. The models trained using the conventional objective $\mathcal{L}$ retain their base names (NPAS and Seq-NPAS), whereas models trained using the proposed rate-aware objective $\mathcal{L}^{++}$ are denoted by NPAS++ and Seq-NPAS++, respectively.

\begin{figure}
    \centering
    \includegraphics[width=\linewidth]{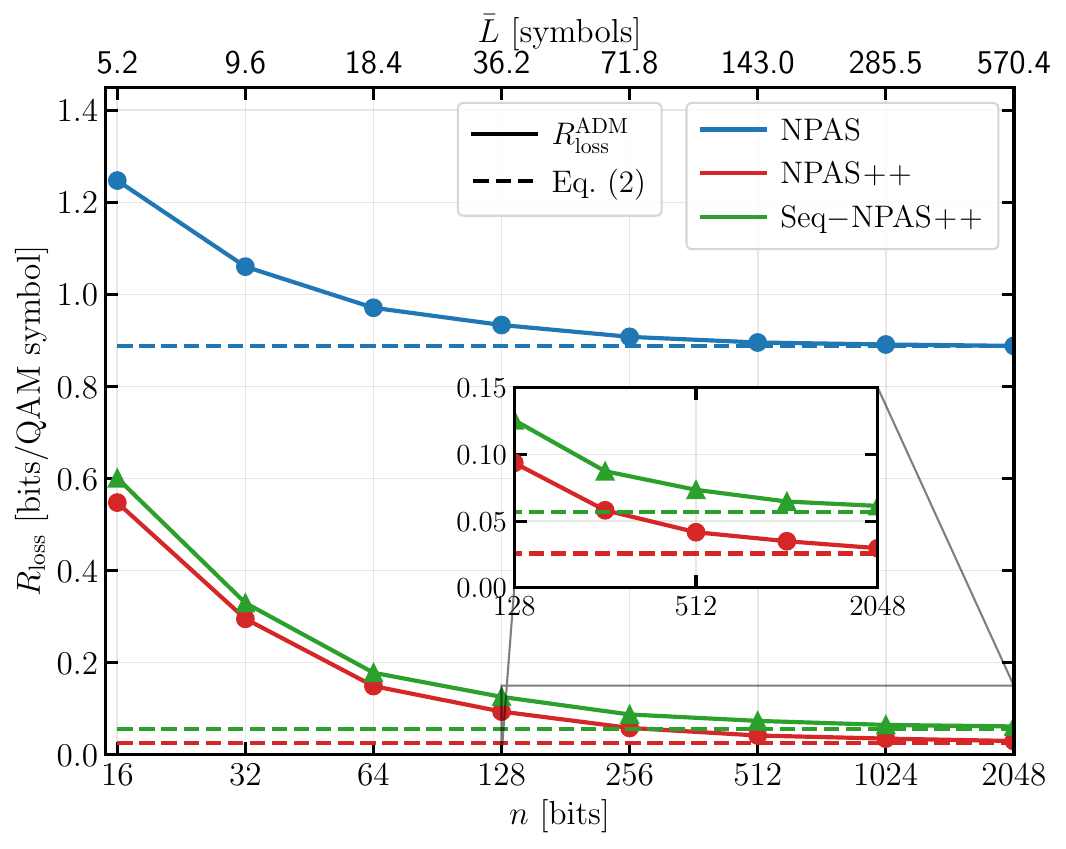}
    \caption{Empirical ADM rate loss versus ADM input length (bottom axis) and average output length (top axis).}
    \label{fig:rateloss}
    \vspace*{2mm}
\end{figure}

We first benchmark the practical rate loss 
\vspace{-2mm}
\begin{equation}
\vspace{-1mm}
R_{\mathrm{loss}}^{\mathrm{ADM}}
=
\frac{1}{N}\sum_{t=1}^{N}H(\ve{b}_t)-\frac{n}{\bar{L}},
\vspace{-1mm}
\end{equation}
achieved by ADM, where $n$ is the number of ADM input bits and $\bar{L}$ is the empirical average output length, 
against the theoretical lower bound in \eqref{eq:rateloss}. 
We consider NPAS with shaping blocklength $4$, trained using both $\mathcal{{L}}$ and $\mathcal{L}^{++}$, and Seq-NPAS++ with $\Nmem\!\!=\!\!15$. Fig.~\ref{fig:rateloss} shows the rate loss versus $n$. As $n$ increases, the empirical ADM rate loss decreases and approaches the theoretical bound. Moreover, joint-distribution learning introduces a non-zero intrinsic rate loss due to symbol dependencies. This loss is substantial when rate loss is ignored during training, whereas it is significantly reduced under the proposed objective, highlighting the importance of rate-aware optimization.

Next, we compare NPAS++ with shaping blocklength $32$, Seq-NPAS++ with $\Nmem\!\!=\!\!15$, uniform signaling, ESS, and ESS combined with sequence selection using the additive-multiplicative metric from \cite{askari2024perturbation}. The ESS shaping rate is set to $1.93$~bits/1D to match the entropy of the optimal MB marginal. Neural models are sampled through ADM with $2048$ input bits. The ESS blocklength and sequence-selection parameters are optimized by grid search to maximize AIR. For the considered link, the best ESS blocklength is $32$, while sequence selection achieves its best performance with selection blocklength $64$ and $16$ candidates. 

\begin{figure}
    \centering
    \includegraphics[width=\linewidth]{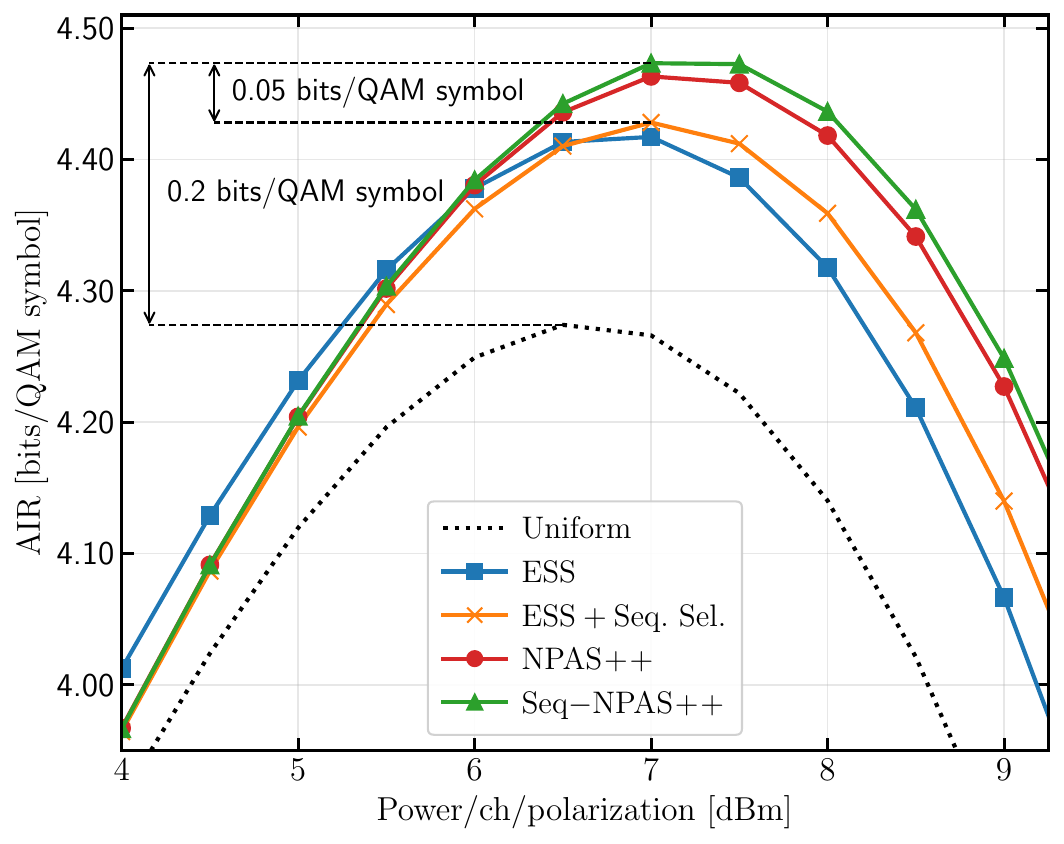}
    \caption{AIR versus launch power.}
    \label{fig:air}
\end{figure}

Fig.~\ref{fig:air} shows the AIR versus launch power. In the low-power regime, nonlinearity-aware schemes underperform ESS because their additional temporal structure introduces rate loss while nonlinear effects remain weak. As launch power increases, joint-distribution learning becomes beneficial. Sequence selection provides negligible gain over ESS, whereas both NPAS++ and Seq-NPAS++ achieve up to $0.05$~bits/2D higher AIR by directly learning the symbol joint distribution while controlling rate loss. 
Overall, Seq-NPAS++ provides the best overall performance.

\section{Discussion and Conclusions}
\label{sec:conclusion}

We present the first practical neural shaping method that outperforms state-of-the-art schemes while explicitly accounting for all implementation losses. Its core innovations are a rate-loss-aware training objective and a sequential autoregressive encoder (Seq-NPAS). By removing block boundaries, Seq-NPAS yields stationary symbol statistics and models long-range dependencies efficiently, enabling easy implementation and close compatibility with ADM. This architecture can learn signal structures well aligned with the channel’s ``language,'' maximizing information transfer to a mismatched receiver in the nonlinear regime. Having demonstrated gains on a single-span fiber link under mismatched training, we expect substantially larger improvements for long-haul links with greater memory and with further training refinements. Importantly, the approach is broadly applicable to other systems affected by nonlinear impairments beyond the fiber Kerr nonlinearity.


\clearpage
\section{Acknowledgements}
This work was supported by Nokia Bell Labs, France; the Mitacs Accelerate International program; the Institute for Computing, Information and Cognitive Systems (ICICS) at UBC; and the Digital Research Alliance of Canada (www.alliancecan.ca).

\defbibnote{myprenote}{%

}
\printbibliography[prenote=myprenote]

\vspace{-4mm}

\end{document}